\documentclass{article}
\usepackage{ijcai24}

\usepackage{soul, color, xcolor}
\usepackage{times}
\usepackage{soul}
\usepackage{url}
\usepackage[hidelinks]{hyperref}
\usepackage[utf8]{inputenc}
\usepackage[small]{caption}
\usepackage{graphicx}
\usepackage{amsmath}
\usepackage{amsthm}
\usepackage{booktabs}
\usepackage{algorithm}
\usepackage{algorithmic}
\usepackage[switch]{lineno}
\usepackage{multirow}
\usepackage{graphicx}
\usepackage{subcaption}
\usepackage{bigstrut}
\usepackage{stfloats}

\newtheorem{theorem}{Theorem}

\title{
A Lightweight U-like Network Utilizing Neural Memory Ordinary Differential Equations for Slimming the Decoder}

\author{
Quansong He$^1$
\and
Xiaojun Yao$^2$\and
Jun Wu$^{1}$\and
Zhang Yi$^{1}$\and
Tao He$^{1,}$\footnote{Corresponding author.}
\affiliations
$^1$College of Computer Science, Sichuan University, Chengdu, China\\
$^2$Department of  Thoracic Surgery, Public Health Clinical Center of Chengdu, China
\emails
hequansong@stu.scu.edu.cn, flyingyao@163.com,
junwu@stu.scu.edu.cn,\\
\{zhangyi, tao\_he\}@scu.edu.cn
}

\begin{document}

\maketitle

\begin{abstract}
In recent years, advanced U-like networks have demonstrated remarkable performance in medical image segmentation tasks. However, their drawbacks, including excessive parameters, high computational complexity, and slow inference speed, pose challenges for practical implementation in scenarios with limited computational resources. Existing lightweight U-like networks have alleviated some of these problems, but they often have pre-designed structures and consist of inseparable modules, limiting their application scenarios. In this paper, we propose three plug-and-play decoders by employing different discretization methods of the neural memory Ordinary Differential Equations (nmODEs). These decoders integrate features at various levels of abstraction by processing information from skip connections and performing numerical operations on upward path. Through experiments on the PH2, ISIC2017, and ISIC2018 datasets, we embed these decoders into different U-like networks, demonstrating their effectiveness in significantly reducing the number of parameters and FLOPs while maintaining performance. In summary, the proposed discretized nmODEs decoders are capable of reducing the number of parameters by about 20\% $\sim$ 50\% and FLOPs by up to 74\%, while possessing the potential to adapt to all U-like networks. Our code is available at
\textcolor{blue}{\url{https://github.com/nayutayuki/Lightweight-nmODE-Decoders-For-U-like-networks}}.
\end{abstract}

\section{Introduction}
Deep learning has become increasingly prominent in the field of paramedicine, providing valuable assistance in disease detection and diagnosis. The UNet \cite{unet} is a significant milestone in demonstrating the efficacy of encoder-decoder Convolutional Neural Networks (CNNs) with skip connections for medical image segmentation. Over time, UNet has become the foundational framework for most notable medical image segmentation methodologies. To achieve higher precision, many studies have introduced complex modules or increased the number of parameters. For example, ResUNet \cite{zhang2018road} combines the UNet architecture with residual connections inspired by ResNet \cite{resnet}. Swin-UNet \cite{swinu} leverages the Swin Transformer \cite{swint} architecture. The TransFuse \cite{transfuse} model employs a dual-path configuration that seamlessly combines CNN and ViT \cite{vit}, enabling it to capture both local and global insights simultaneously.

\begin{figure}
\centering
\begin{minipage}{0.23\textwidth}
	\includegraphics[width=\textwidth]{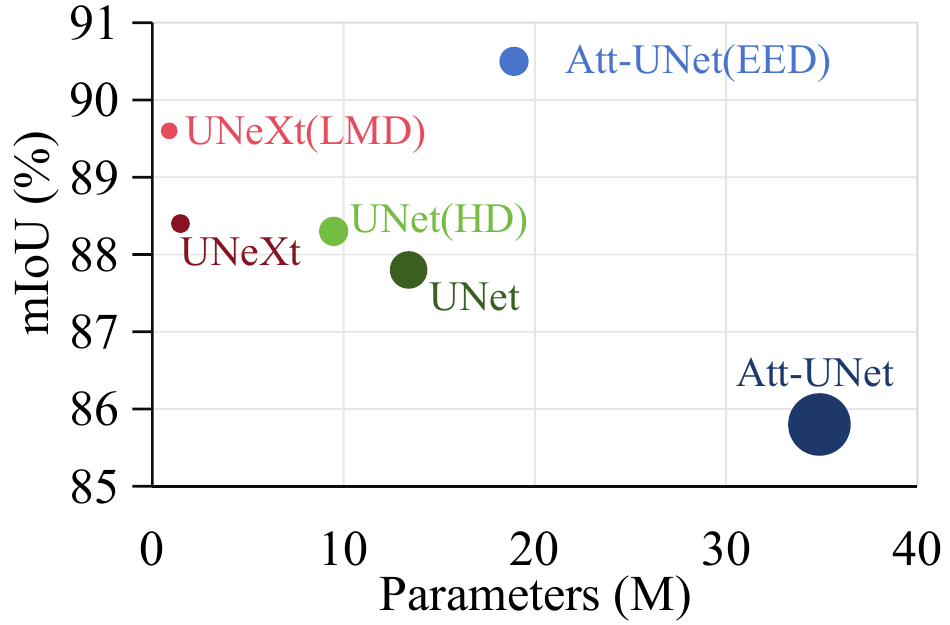}\subcaption{mIoU on PH2.}
\end{minipage}
\begin{minipage}{0.23\textwidth}
	\includegraphics[width=\textwidth]{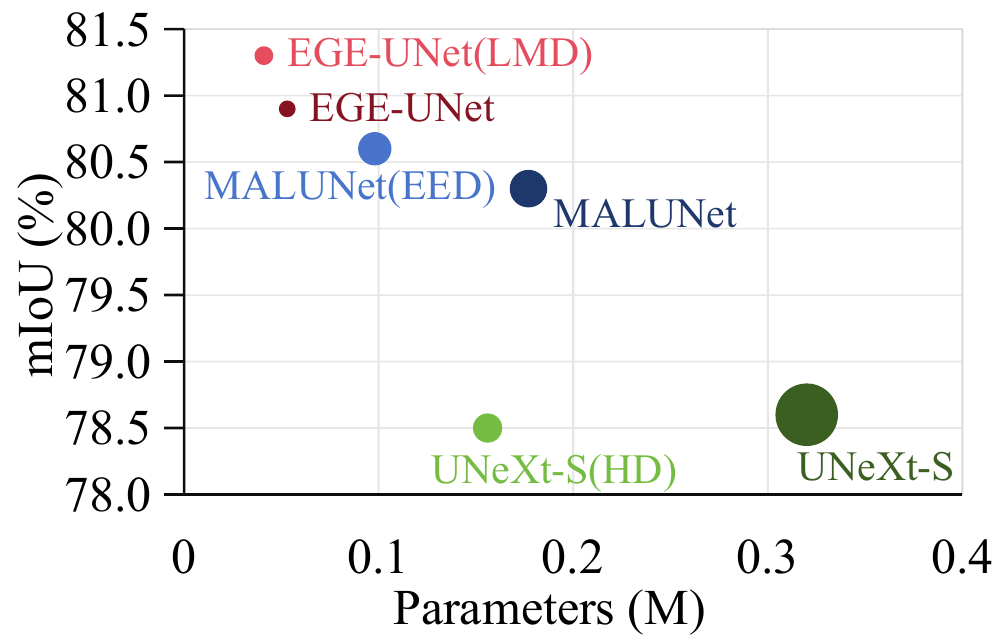}\subcaption{mIoU on ISIC2018.}
\end{minipage}
\caption{Visualization of the mIoU results on the PH2 and ISIC2018 datasets. X-axis
corresponds to the number of parameters (lower the better). Y-axis represents the mIoU (higher the better). The circle's area is proportional to FLOPs (smaller the better).}
\label{fig1:env}
\end{figure}

Existing U-like networks have improved the performance of UNet, but most of them come with an increased computational cost. This can pose challenges when attempting to deploy them in practical application scenarios with limited computing resources. To address this issue, researchers have ventured into the research direction that goes beyond the pursuit of model performance, focusing on model lightweighting. In recent developments, UNeXt \cite{unext} combined UNet and MLP \cite{mlp}, presenting a lightweight architecture that achieves remarkable performance while reducing both parameter count and computational requirements. MALUNet \cite{malunet} has successfully reduced model size by decreasing the number of model channels and incorporating multiple attention modules. The Efficient Group Enhanced UNet (EGE-UNet) \cite{ege} reduced the number of parameters and computational complexity through the integration of enhanced attention mechanisms and feature fusion modules.

While the aforementioned methods have demonstrated impressive abilities in reducing model parameters and computational complexity, they often face challenges when it comes to adaptability to existing frameworks and lack universality. 
To achieve lightweighting, most of these networks heavily reduce the number of channels in their network structure, resulting in a trade-off between network performance and the reduction of parameters and computations. However, these networks often pursue lightweighting and network performance as separate objectives, lacking a comprehensive solution. In order to bridge the gap between network performance and computational complexity, this paper aims to explore a universal method that has the potential to adapt to all U-like networks.

In this paper, we integrate the neural memory Ordinary Differential Equations (nmODEs) \cite{nmODE} decoders into U-like networks using three different discretization methods: \emph{\textbf{explicit Euler's method}}, \emph{\textbf{Heun's method}}, and \emph{\textbf{linear multistep method}}. We then assess their effectiveness on multiple datasets. The application of the nmODEs decoders across different UNet variants showcases the versatility of our proposed solution. Moreover, experimental results substantiate that the nmODEs decoders effectively reduce parameters and computational complexity, without compromising the performance of the original model, and in some cases even slightly improving it. In summary, the proposed discretized nmODEs decoders can decrease the parameter count by approximately 20\% $\sim$ 50\% and the FLOPs by up to 74\%.

\section{Related Work}
To delve into the integration of U-like networks with the theme of nmODEs, this section will first introduce several classic UNet networks and methods related to nmODEs.

\subsection{UNet and Its Variants}
UNet \cite{unet} was the cornerstone of all UNet variants for semantic image segmentation. It has a U-like structure with a contracting path for context extraction and an expanding path for precise localization. Utilizing skip connections, UNet effectively preserved fine feature details. Attention UNet (Att-UNet) \cite{attu} incorporated an attention mechanism into the UNet, significantly improving its efficacy in extracting image features. UNeXt \cite{unext}, being the pioneering lightweight medical image segmentation network that integrates convolutional and multilayer perceptrons, had shown remarkable success in reducing network parameters and computational workload. MALUNet  \cite{malunet} demonstrated outstanding performance in skin cancer segmentation by incorporating attention mechanisms and deep separable convolution into UNet through carefully designed modules. Building upon MALUNet, EGE-UNet \cite{ege} further refined the attention mechanism and introduces new feature fusion modules, ensuring network performance while significantly reducing network complexity. EGE-UNet surpassed many large-scale networks in terms of both performance and efficiency. 

\subsection{Neural Ordinary Differential Equations}
Ordinary Differential Equation (ODE) systems, recognized as a distinctive class of dynamical systems, have long been subject to extensive exploration and empirical investigation in the realms of mathematics and physics. Neural ODEs (NODEs) \cite{node} provided mathematical principles that elucidate ResNet, transforming it from an enigmatic black-box network into a comprehensible framework. They presented a novel approach to conceptualize neural networks as representations of ODEs. NODEs established a foundation for the unification of neural networks and ODEs. One notable advantage of NODEs is their ability to eliminate the storage of intermediate quantities during the forward propagation, resulting in a substantial reduction in parameters and computational overhead. Nevertheless, there are certain limitations in mapping data through NODEs. For instance, they face challenges in representing mappings like $g(1) = -1$, $g(-1) = 1$. This limitation arises from the fact that NODEs models utilizing data inputs as initial values can only learn features within the same topological space as the input data \cite{dupont}. Furthermore, when modeling problems with differential equations, creating a dynamical system, it has been demonstrated that attractors in dynamical systems are believed to be linked to memory capacity \cite{attractor1} \cite{attractor2}. However, conventional NODEs lack the capability to effectively leverage the memory capacity provided by attractors. 

\vspace{-0.25mm}
The nmODEs \cite{nmODE} are specialized variants of NODEs designed to overcome the limitations inherent in traditional NODEs and harness the full memory capabilities offered by dynamical systems. It enhanced the neural network's nonlinear expression capability by employing implicit mapping and utilizing nonlinear activation functions. Simulating the dynamical system governing neocortical neuronal memory, nmODEs introduced a distinct feature—a clear dynamical characterization achieved through the segregation of learning neurons and memory neurons. Unlike previous NODEs approaches, nmODEs treated input data as external parameters rather than utilizing them as initial values for ODEs. By separating the neuron's function into learning and memory components, learning exclusively occurs in the learning part, while the memory part maps the input to its global attractor, establishing a mapping from input space to memory space. The nmODEs have found successful applications in various segmentation tasks. For instance, the nmPLS-Net \cite{nmpls} leveraged the robust nonlinear representation and memory capabilities of nmODEs to construct an edge segmentation-based decoding network. This approach has enabled accurate lung lobe segmentation. Additionally, the integration of nmODEs into UNet, employing a straightforward discretization method, has demonstrated promising outcomes in tasks such as diabetic kidney segmentation \cite{nmunet} and liver segmentation \cite{nmdecoder}. \cite{robustness} enhanced the robustness of medical image segmentation using nmODEs and yielded favorable outcomes. These applications showcase the versatility and effectiveness of nmODEs across different medical image segmentation challenges.

\section{Methods}
In this section, we leverage the discretization methods of nmODEs to construct a lightweight U-like network architecture. nmODEs have the nature to be adaptive to the decoders of U-like networks. Therefore, in Section \ref{3.1}, we will introduce the U-like networks with the discretized nmODEs decoders. These U-like networks utilize skip connections to extract low-level features and inject them into internal states for feature aggregation, resulting in a series of low parameter nmODEs decoders. In Section \ref{3.2}, we will introduce three discretized ODE solvers to approximately calculate the output of the nmODEs decoders.

\begin{figure}
\centering
\includegraphics[width=1\linewidth]{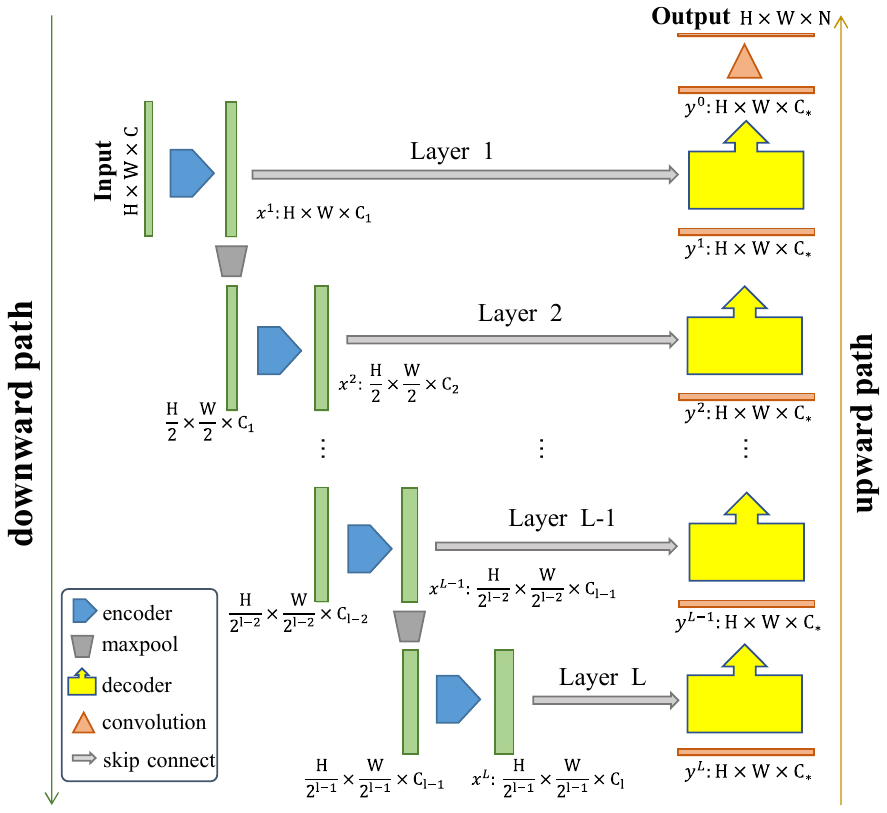}
\caption{The U-like networks (with the proposed discretized nmODEs decoders). The upward path of the nmODEs decoders share the same parameters among layers without upsampling operations.}
\label{fig:arch}
\end{figure}

\subsection{The U-like Networks with nmODEs Decoders}
\label{3.1}
\begin{figure}[h]
\centering
\begin{minipage}{0.22\textwidth}
	\includegraphics[width=\textwidth]{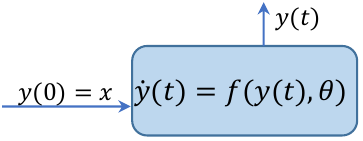}\subcaption{NODE block.}
\end{minipage}\hspace{5pt}
\begin{minipage}{0.22\textwidth}
	\includegraphics[width=\textwidth]{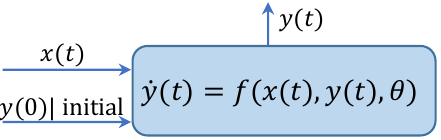}\subcaption{nmODE block.}
\end{minipage}
\caption{Comparison between general NODE and nmODE.}
\label{fig3:env}
\end{figure}

To begin with, let's revisit the disparity in inputs 
between conventional NODE and nmODE. Their structures are delineated in Fig. 3(a) and Fig. 3(b), respectively. In the typical NODEs configuration, the initial value $y(0)$ is derived from the data itself, and the output represents the numerical solution of the ODE. In contrast, within the nmODEs architecture, the initial value $y(0)$ is set to a random value, e.g., $0$. Data served as an sequential external input $x(t)$. nmODEs take two inputs, aligning seamlessly with the decoders of U-like networks. Information from the skip connections serves as the sequential external inputs to the nmODEs decoders, while information from the upward path serves as $y(t)$, enabling the full utilization of nmODEs. This is exactly the reason why we chose nmODEs. The differential equation for nmODEs is formulated as follows: 

\begin{equation}\label{eqn:nmODE}
    \Dot{y}(t)=-y(t) + f\left(y(t)+g(x(t),\theta_t)\right) .
\end{equation}

For $t \geq 0$, where $y(t) \in R^n $ represents the network's state, $x(t) \in R^m $ stands for external input, and $\theta_{t}$ means the parameters of the skip connections. We use the nmODEs to formulate the decoders of the U-like networks, resulting in an modified U-like network as illustrated in Fig. \ref{fig:arch}. This network applies parameterized computation in the skip connections, which is denoted as $g(x(t),\theta_t)$. The duty of the skip connections is to transform the low-level features to the fixed-size high-level feature maps. The U-like network with nmODEs decoders performs parameter-less feature aggregation. Given an $L$-layer U-like network using nmODEs decoders, define the output of the network encoder, which is the skip-connected input of the decoder, as $x^l$, and denote the upward path input of the decoder as $y^l$, $1 \leq l \leq L$.Initializing $x(0)=x^L$ and $y(0)=y^L=0$, the target of the nmODEs decoders is to obtain the numerical solution at the point $(\tau, x(\tau)=x^1)$ along the trajectory of the Eq. (\ref{eqn:nmODE}). In the rest of this section, we will introduce three discretized ODE solvers, e.g. explicit Euler's method, Heun's method, and linear multistep method.

For simplicity in subsequent discussions, let's denote Eq. (\ref{eqn:nmODE}) as $F(t,y)$ , the function $F$ remains continuous and adheres to certain Lipschitz conditions, ensuring the solution's existence and uniqueness. Given an initial value $y_{0}$, the trajectory traced by Eq. (\ref{eqn:nmODE}) starting from $y_{0}$ is denoted as $y_{t}$, encompassing all instances where $ t \geq 0$. A vector $y^*$ is defined as an equilibrium point of the NODEs if it satisfies the equation $F(t^*,y^*) = 0$. An equilibrium point $y^*$ earns the designation of a global attractor when, for any given $y_{0}$, the corresponding trajectory $y_{t}$ converges towards $y^*$ as $t \rightarrow \infty$. With the existence of global attractor within nmODEs, the establishment of a favorable nonlinear mapping from $x$ to $y^*$ becomes feasible.

\subsection{Discretization Methods}
\label{3.2}
In the modified lightweight U-like networks, the upward path is parameterless. Only a small number of parameters are responsible for information integration of skip connections, matching the feature sizes at the upward path. We discretize the nmODEs decoders in three different methods and replace the original decoders of U-like networks. The structure inside the decoders varies with the discretization method, with specific reference to the mathematical derivation. All three discretization methods address initial value problems (IVPs) for nmODEs, sharing common steps. 

\begin{figure}[h]
\centering
\includegraphics[width=1\linewidth]{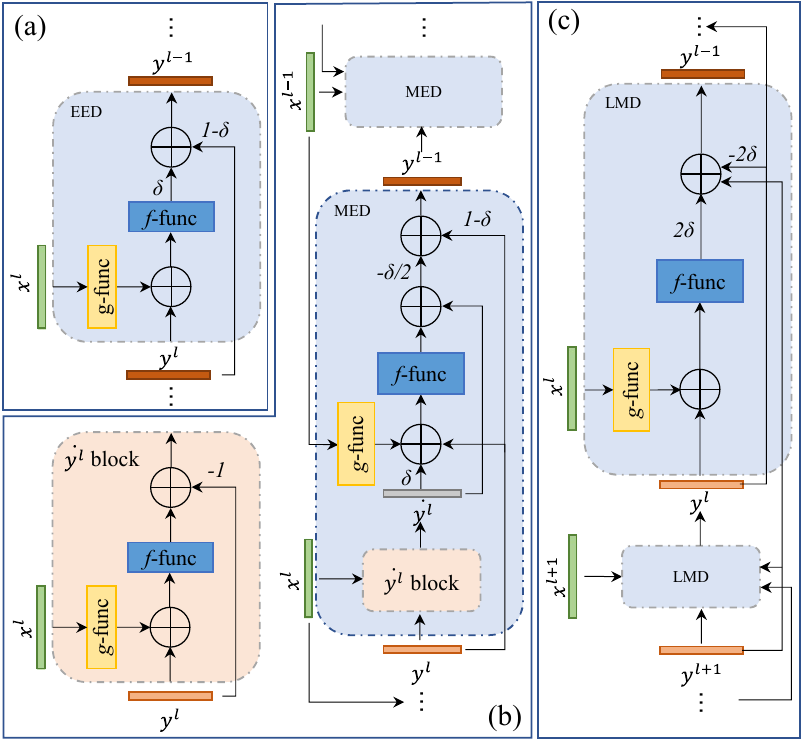}
\caption{(a) explicit Euler's method discretized nmODEs decoder (EED). (b) Heun's method discretized nmODEs decoder (HD). (c) linear multistep method discretized nmODEs decoder (LMD). The blue part of each figure shows the complete decoder of the current layer, and the orange $y^l$ and red $y^{l-1}$ are used as the upward path inputs and outputs of the current layer, respectively.}
\label{fig4:env}
\end{figure}

\begin{theorem}
	\label{theo1}
	\textbf{Explicit Euler's Method \cite{euler1845institutionum}.} Given the derivative $\Dot{y}(t)=F(t, y(t))$, choose a value $\delta$ for the size of every step along t-axis and set $t_{n+1}=t_n+\delta$. The $y_{n+1}$ from $y_n$ and $t_n$ is 
	\begin{equation}\label{eqn:euler}
		y_{n+1} = y_n+\delta\cdot \Dot{y}(t_n) = y_n+\delta\cdot F(t_n, y_n),
	\end{equation}
	where $y_n$ is an approximate solution at time $t_n$, i.e., $y_n \approx y(t_n)$. $y_{n+1}$ is an explicit function of $y_i$ for $i\leq n$.
\end{theorem}

The explicit Euler's method, defined in \textbf{Theorem} \ref{theo1}, is the most straightforward in concept and also the most convenient discretization approach to apply. Let $t=t_n$, we can derive from the Eq. (\ref{eqn:nmODE}) and (\ref{eqn:euler})
that
\begin{equation}\label{eqn:dis_euler}
    \begin{aligned}
     y_{n+1} &= y_n + \delta\cdot\left(
     -y_n + f(y_n+g(x_n,\theta_n))
     \right) \\
     &= (1-\delta)\cdot y_n + \delta\cdot f(y_n+g(x_n,\theta_n)).
     \end{aligned}
\end{equation}

Eq. (\ref{eqn:dis_euler}) is formulated for continuous values, and adapting it to discretized neural networks necessitates a conversion. In the context of a U-like network structure, where the value $y^l$ in the lowest layer serves as the initial value, the initial value problem is solved during the operation of the upward path. This process results in a discrete form, with the number of solution steps being inversely proportional to the number of network layers:

\begin{equation}\label{eqn:dis_euler_net}
     y^{l-1}= (1-\delta)\cdot y^l + \delta\cdot f(y^l+g(x^l,\theta^l)) .
\end{equation}
In the context of applying U-like networks, the total number of layers is L. $l$ is the number of layers of the current operation, $1 \leq l \leq L$. $y^l$ in Eq. (\ref{eqn:dis_euler_net}) represents the input of the $l$-th layer of the upward path, serving as the initial value for solving $y^{l-1}$. Here, $x^l$ denotes the information transmitted through the skip connection, and $\delta$ is $1/L$. $\theta^l$ means the parameters for skip connection in $l$-th layer. Based on Eq. (\ref{eqn:dis_euler_net}) we show the internal realization of the decoder as Fig. 4 (a).

\begin{theorem}
	\label{theo2}
	\textbf{Heun’s method \cite{heun}.} Given the derivative $\Dot{y}(t)=F(t, y(t))$, choose a value $\delta$ for the size of every step along t-axis and set $t_{n+1}=t_n+\delta$. 
	First calculate the intermediate value $\overline{y_{n+1}}$  and then the final approximation $y_{n+1}$ at the next integration point as follows:
	\begin{equation}\label{eqn:modeuler}
	\begin{aligned}
     \overline{y_{n+1}} &= y_n+\delta\cdot \dot{y}(t_n) = y_n+\delta\cdot F(t_n, y_n)\\
     y_{n+1}&= y_n + \frac{\delta}{2}\cdot (\dot{y}(t_n)+\overline{\dot{y}(t_{n+1})}),
    \end{aligned}
	\end{equation}
	where $y_n$ is an approximate solution at time $t_n$, i.e., $y_n \approx y(t_n)$. $\overline{y_{n+1}}$ is the predicted median, $y_{n+1}$ is the final result corrected for $\overline{y_{n+1}}$.
\end{theorem}

The Heun’s method, defined in \textbf{Theorem} \ref{theo2}, replaces the derivative at the point $y_n$ in the explicit Euler's method with the average of derivatives at $y_n$ and $y_{n+1}$. Therefore, from a mathematical perspective, this method enhances accuracy compared to the explicit Euler's method. Let $t=t_n$, we can derive from the Eq. (\ref{eqn:nmODE}) and (\ref{eqn:modeuler}) that
    \begin{multline}\label{eqn:modeuler_net}
        \begin{aligned}
     y_{n+1}&= y_n + \frac{\delta}{2}\cdot (\Dot{y}(t_n)+F(t_{n+1},\overline{y_{n+1}}))\\
            &= y_n + \frac{\delta}{2}\cdot (\Dot{y}(t_n)-\overline{y_{n+1}}+\\
            & \qquad \qquad  f(\overline{y_{n+1}}+g(x_{n+1}.\theta_{n+1})))\\
            &= y_n + \frac{\delta}{2}\cdot (\Dot{y}(t_n)-y_n-\delta\cdot \Dot{y}(t_n)+\\
            & \qquad \qquad   f(y_n+\delta\cdot \Dot{y}(t_n) + g(x_{n+1}.\theta_{n+1})))\\
            &= (1-\frac{\delta}{2})\cdot y_n + \frac{\delta}{2}\cdot \big[(1-\delta)\cdot \Dot{y}(t_n)+\\
            & \qquad \qquad   f(y_n+\delta\cdot \Dot{y}(t_n)+g(x_{n+1},\theta_{n+1}))\big].
    \end{aligned}
    \end{multline}\label{eqn:dis_modeuler}
Since we design reusable modules for $\Dot{y}(t_n)$, there is no more substitution of expansion in Eq. (\ref{eqn:modeuler_net}). Similarly, we rewrite the continuous numerical form into a discretized network form to obtain the following equation:
    \begin{multline}
     y^{l-1}= (1-\frac{\delta}{2})\cdot y^l + \frac{\delta}{2}\cdot \big[(1-\delta)\cdot \Dot{y}^l+\\
             \quad f(y^l+\delta\cdot \Dot{y}^l+g(x^{l-1},\theta^{l-1}))\big].
    \end{multline}\label{eqn:HD}
The decoder structure based on this construction is illustrated in Fig. 4 (b). It is worth noting that the Heun's method requires two layers of skip connection information as input $x^l$ and $x^{l-1}$: the current layer and the next layer. Which means $2 \leq l \leq L$. Therefore, the decoder of the layer $1$ applies an EED.

\begin{theorem}
	\label{theo3}
	\textbf{Linear Multistep Method \cite{adams}.} Given the derivative $\Dot{y}(t)=F(t, y(t))$, choose a value $\delta$ for the size of every step along t-axis and set $t_{n+1}=t_n+\delta$ as follows:
	\begin{equation}\label{eqn:linear_multistep}
	\begin{aligned}
     y_{n+1} &= y_{n-1} + 2\cdot\delta\cdot\dot{y}(t_n) \\
     &= y_{n-1} + 2\cdot\delta\cdot F(t_n, y_n),
    \end{aligned}
	\end{equation}
	where $y_n$ is an approximate solution at time $t_n$, i.e., $y_n \approx y(t_n)$. $y_{n+1}$ is the final result associated with both $y_{n}$ and $y_{n-1}$.
\end{theorem}

The linear multistep method, defined in \textbf{Theorem} \ref{theo3}, involves a linear combination of the derivatives of multiple selected points, the most commonly used number of selected points is two, and the calculation of the unknown points is performed by the information of the known two points, this linear multistep method is also known as the Eulerian two-step method. Let $t=t_n$, we can derive from the Eq. (\ref{eqn:nmODE}) and Eq. (\ref{eqn:linear_multistep}) that
\begin{equation}\label{eqn:linear_euler}
    \begin{aligned}
     y_{n+1} &= y_{n-1} + 2\cdot\delta\cdot\left(
     -y_n + f(y_n+g(x_n,\theta_n))
     \right) \\
     &= y_{n-1} - 2\cdot\delta\cdot y_n + 2\cdot\delta\cdot f(y_n+g(x_n,\theta_n)).
     \end{aligned}
\end{equation}

Rewrite it in discretized form:
\begin{equation}\label{eqn:linear_euler_net}
    y^{l-1} = y^{l+1} - 2\cdot\delta\cdot y^l + 2\cdot\delta\cdot f(y^l+g(x^l,\theta^l)) .
\end{equation}  

The internal structure of the decoder designed based on Eq. (\ref{eqn:linear_euler_net}) is shown in Fig. 4 (c). It is worth noting that the linear multistep method requires two layers of upward path information as input $y^l$ and $y^{l+1}$, the current layer and the previous layer. Which means $1 \leq l \leq L-1$, so the decoder of the layer $L$ applies a EED.

\subsubsection{Function Selection \& Initial Value Problem}

In the network framework we devised, the upward path is parameter-free, meaning the $f$-function has no parameters to alter the shape of input $y^l$. Consequently, the responsibility of reshaping $x^l$ to match $y^l$ falls on the $g$-function. Therefore, the $g$-function includes essential components like the up-sampling function to modify height and width, and the convolution operation to adjust the number of channels. Moreover, we have the flexibility to incorporate parameter-free operations, such as batch normalization and activation functions, within both the $g$ and $f$ functions. In this paper, the internal implementations of the $g$-function include, in sequence, the convolution, upsampling, and activation functions, while the $f$-function exclusively comprises batch normalization. For the initialization of $y(0)$, we set it to a zero matrix, maintaining consistency with the initial input width and height. However, there is some flexibility in determining the channels of $y(0)$. Typically, the number of channels is chosen to be consistent with the original input, but considerations are also given to the number of groups to facilitate grouped convolution operations. As demonstrated in the experimental section, augmenting the number of channels enhances performance to some extent, albeit accompanied by an escalation in the number of model parameters and FLOPs.

\section{Experiments}
\subsection{Datasets}
\textbf{PH2.} The PH2\footnote{https://www.fc.up.pt/addi/ph2\%20database.html} dataset is a compilation of 200 dermoscopic images focused on dermatology, specializing in melanocytic lesions. It enables research in skin lesion segmentation and classification, with a particular emphasis on nevus and melanoma.\\
\textbf{ISIC2017.} The ISIC2017\footnote{https://challenge.isic-archive.com/data/\#2017} dataset is a comprehensive compilation of 2150 dermoscopic images of skin lesions. It serves as a valuable resource for advancements in dermatology and computer-aided diagnosis. The dataset covers a wide range of skin conditions, including both benign and malignant lesions. It is primarily designed for tasks related to melanoma detection and skin cancer classification.\\
\textbf{ISIC2018.} The ISIC2018\footnote{https://challenge.isic-archive.com/data/\#2018} dataset is a valuable resource in dermatological image analysis, focusing on melanoma detection and skin cancer classification. It consists of approximately 2700 diverse dermoscopic images.

\subsection{Implementation Details}
In our experiments, we replaced the decoders of UNet, Att-UNet, MALUNet, EGE-UNet and UNeXt with nmODEs decoders using various discretization methods. All experiments were executed on a single RTX 4090 GPU using PyTorch\footnote{https://pytorch.org/get-started/locally/}. Each dataset was randomly split into training and testing sets with a 7:3 ratio, and images were normalized and resized to 256 $\times$ 256 for consistency. Data augmentation techniques, including horizontal flipping, vertical flipping, and random rotation, were applied. The optimizer AdamW \cite{adamw} was used, along with the CosineAnnealingLR \cite{cos} scheduler, setting the maximum iteration count to 50 and the minimum learning rate to 1e-5. Training spanned 300 epochs with a batch size of 8. Mean Intersection over Union (mIoU) and Dice similarity score (DSC) were employed as evaluation metrics. Experiments were repeated five times, and results were reported as mean and standard deviation for each dataset.

\subsection{Comparative Results}

\begin{table*}[t]
  \centering
    \begin{tabular}{c|c|cc|cc}
    \hline
    
    \hline
    \multicolumn{1}{c|}{\textbf{Dataset}} & \textbf{Model} & \textbf{Params(M)} & \textbf{GFLOPs} & \textbf{mIoU}  & \textbf{DSC} \bigstrut\\
    \hline
    
    \hline
    \multirow{9}[2]{*}{PH2} 
          & Swin-UNet \cite{swinu}$^\ast$  & 25.86 & 5.86  & 0.872  & 0.927  \bigstrut\\
          & UNet++ \cite{unet++}$^\ast$ & 25.66 & 28.77 & 0.883  & 0.936  \\
          & SegNetr \cite{segnetr}$^\ast$ & 12.26 & 10.18 & \textbf{0.905}  & 0.948  \bigstrut\\
          & Att-UNet \cite{attu}$^\dag$ & 34.88 & 66.63 & 0.858  & 0.903  \\
          & \textbf{Att-UNet (EED)} & 18.91 & 17.34 & \textbf{0.905±0.003} & \textbf{0.950±0.002} \bigstrut\\
          & UNeXt \cite{unext}$^\ast$ & 1.47  & 0.57  & 0.884  & 0.936  \\
          & \textbf{UNeXt (LMD)} & \textbf{0.88} & \textbf{0.37} & 0.896±0.003 & 0.945±0.002 \bigstrut\\
          & UNet \cite{unet}$^\ast$ & 13.40  & 31.12 & 0.878  & 0.919  \\
          & UNet (HD) & 9.48 & 15.11 & 0.883±0.005  & 0.938±0.003  \bigstrut\\
    \hline
    
    \hline
    \multirow{9}[2]{*}{ISIC2017} & FAT-Net \cite{fatnet}$^\dag$ & 88.87 & 24.63 & 0.765  & 0.850  \bigstrut\\
          & MobileViTv2 \cite{mobilevitv2}$^\star$ & 1.87  & 0.70  & 0.787  & 0.881  \\
          & TransFuse \cite{transfuse}$^\star$ & 26.16  & 11.50  & 0.792  & 0.884  \bigstrut\\
          & EGE-UNet \cite{ege}$^\star$ & 0.053 & \textbf{0.072} & \textbf{0.798}  & \textbf{0.888}  \\
          & \textbf{EGE-UNet (EED)} & \textbf{0.041} & 0.077 & \textbf{0.797±0.001} & \textbf{0.887±0.001} \bigstrut\\
          & UNet \cite{unet}$^\star$ & 13.40  & 31.12 & 0.770  & 0.870  \\
          & UNet (LMD)& 9.45 & 14.89 & 0.773±0.001 & 0.872±0.001 \bigstrut\\
          & MALUNet \cite{malunet}$^\star$ & 0.177 & 0.085 & 0.788  & 0.881  \\
          & MALUNet (HD)& 0.101 & 0.095 & 0.786±0.003 & 0.880±0.002 \bigstrut\\
    \hline
    
    \hline
    \multirow{9}[2]{*}{ISIC2018} & UTNetV2 \cite{utnetv2}$^\star$& 12.80  & 15.50  & 0.790  & 0.883  \bigstrut\\
          & SANet \cite{sanet}$^\star$& 23.9  & 5.96  & 0.795  & 0.886  \\
          & TransFuse \cite{transfuse}$^\star$ & 26.16  & 11.50  & 0.806  & 0.893  \bigstrut\\
          & MALUNet \cite{malunet}$^\star$ & 0.177 & 0.085 & 0.803  & 0.890  \\
          & MALUNet (EED) &0.098 & 0.082 & 0.806±0.002 & 0.893±0.001 \bigstrut\\
          & EGE-UNet \cite{ege}$^\star$ & 0.053 & \textbf{0.072} & 0.809  & 0.895  \\
          & \textbf{EGE-UNet (LMD)} & \textbf{0.041} & 0.077 & \textbf{0.813±0.005} & \textbf{0.897±0.003} \bigstrut\\
          & UNeXt-S \cite{unext}$^\star$& 0.32  & 0.10   & 0.791  & 0.883  \\
          & UNeXt-S (HD) & 0.156 & 0.079 & 0.784±0.003 & 0.879±0.002 \bigstrut\\
    \hline
    
    \hline
    \end{tabular}%
  \caption{Experimental comparison results. $\ast$, $\dag$, and $\star$ means the experimental results are from existing works, where $\ast$ are from \protect\cite{segnetr}, $\dag$ are from \protect\cite{fatnet}, $\star$ are from \protect\cite{ege}, respectively. Bold in the table emphasizes superior performance or very lightweight data presentation.}
  \label{tab:addlabel1}%
\end{table*}%

We chose five prevalent networks from the existing U-like networks, each with documented experimental results on relevant datasets. Subsequently, we replaced their decoders with nmODEs decoders employing different discretizations. The comparison with the original networks and other major models is illustrated in Table 1.
The experimental results clearly demonstrates the undeniable effectiveness of our proposed nmODEs decoders in terms of parameter and FLOPs reduction. For instance, the Att-UNet, which initially boasts the highest number of parameters, sees a reduction of approximately 46\% in parameters and 74\% in FLOPs after applying EED. Similarly, the classical UNet experiences a reduction of about 30\% in parameters and 40\% in FLOPs with the application of HD. Even in the case of lightweight networks, a consistent reduction of around 50\% in parameters and 20\%-30\% in FLOPs is observed after integrating the nmODEs decoders.

In terms of performance, the networks with the nmODEs decoders generally outperform the original networks, while those without performance improvement largely maintain the level of the original networks. On the PH2 dataset, both mIoU and DSC of Att-UNet with EED applied improve by about 5 percentage points over the original network, reaching the state-of-the-art. Similarly, mIoU and DSC of UNet with LMD applied improve by 2.7 and 1.6 percentage points over the original network, also reaching the state-of-the-art, as demonstrated in the ablation experiment section. On the ISIC2017 dataset, the networks employing nmODEs decoders largely retain the performance of the original networks. On the ISIC2018 dataset, EGE-UNet with LMD applied outperforms the original network and reaches the state-of-the-art level. 

All three of our proposed decoders effectively reduce the number of parameters and FLOPs without compromising the original network's performance. However, among these three decoders, HD, with one additional g-function compared to EED and LMD, incurs more parameters and FLOPs. Surprisingly, its performance is not as strong as that of the other two decoders, raising questions that warrant further investigation.

\subsection{Representative Results}
Figures 5 illustrates the qualitative segmentation results obtained by replacing the decoders of MALUNet, EGE-UNet, Att-UNet and UNet with the nmODEs decoders. These comparisons are made against the original network using partially representative images. In each figure, the last two columns highlight that the mask predicted by the network with the nmODEs decoders is closer to the ground truth than the mask predicted by the original network.

\begin{figure*}[t]
\centering
\begin{minipage}{0.23\textwidth}
	\includegraphics[width=\textwidth]{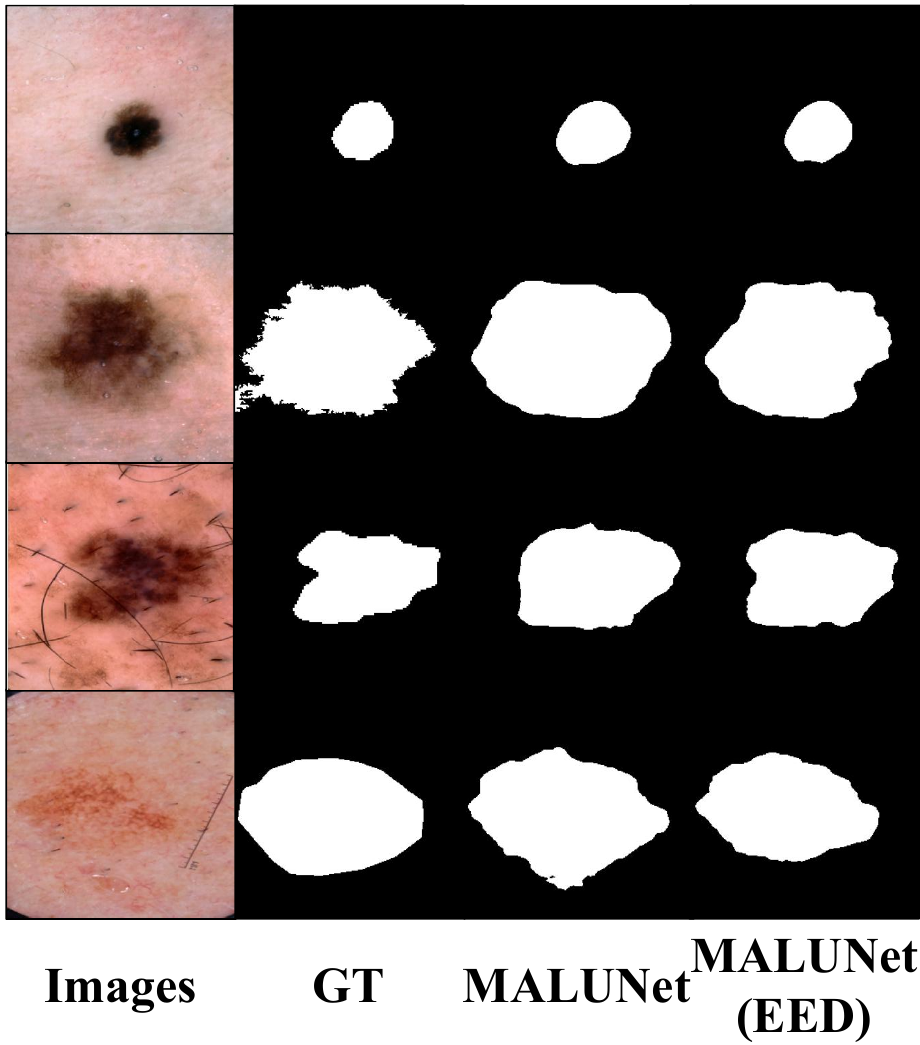}\subcaption{Visualization on the ISIC2018}
\end{minipage}
\begin{minipage}{0.23\textwidth}
	\includegraphics[width=\textwidth]{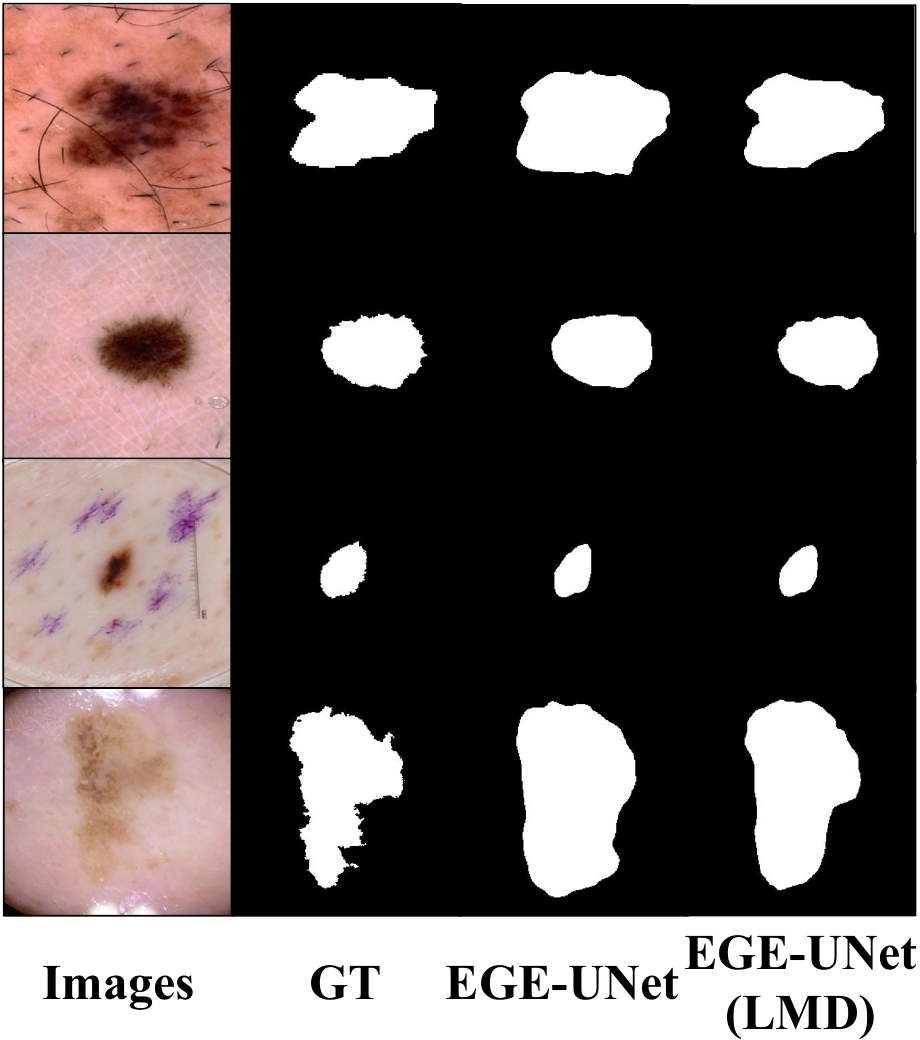}\subcaption{Visualization on the ISIC2018}
\end{minipage}
\begin{minipage}{0.23\textwidth}
	\includegraphics[width=\textwidth]{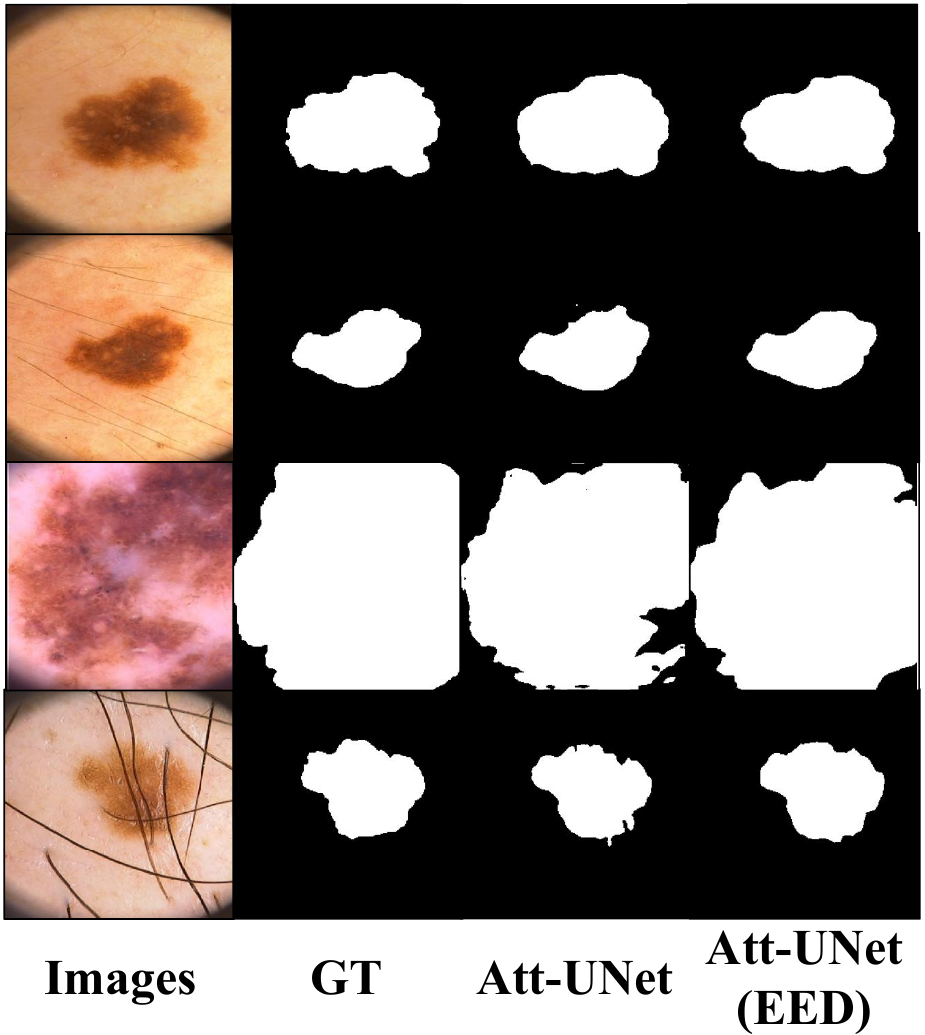}\subcaption{Visualization
on the\\ PH2}
\end{minipage}
\begin{minipage}{0.23\textwidth}
	\includegraphics[width=\textwidth]{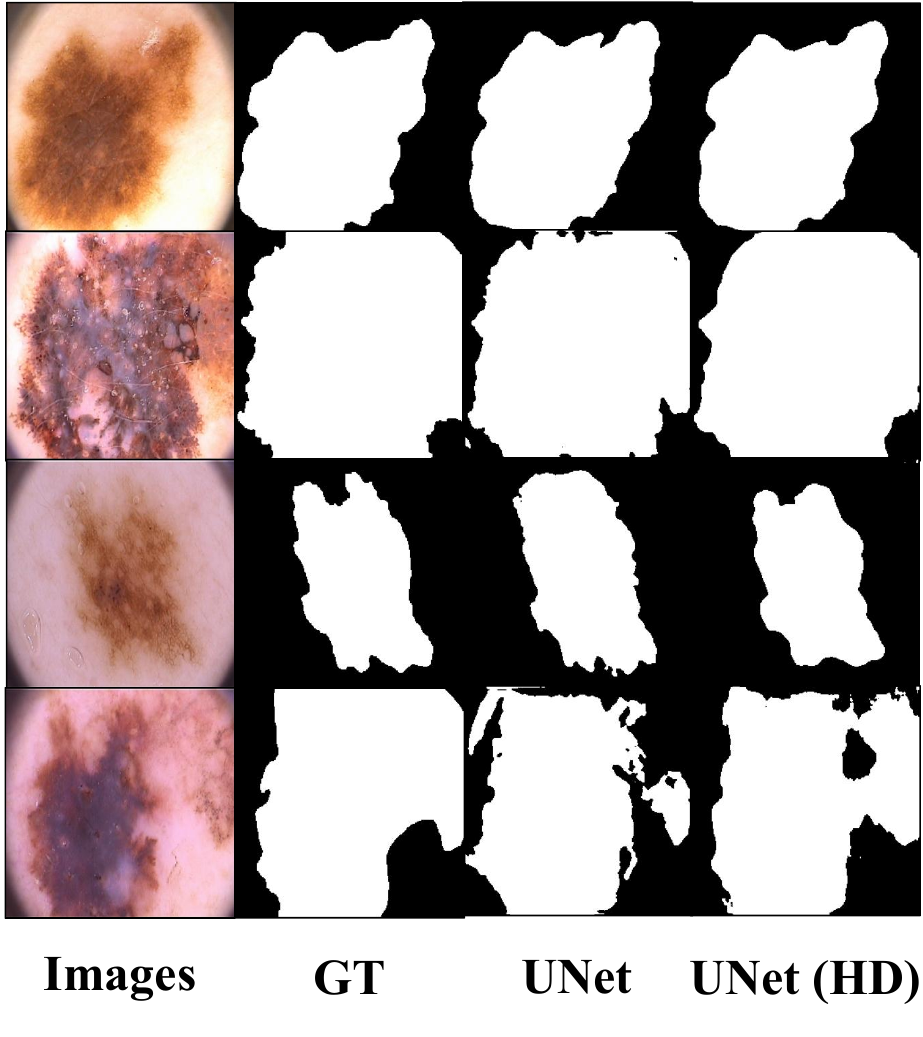}\subcaption{Visualization on the ISIC2017}
\end{minipage}
\caption{Comparison of segmentation results between original networks and networks utilizing nmODEs decoders. Fig (a), (b), (c) and (d) depict the segmentation effects comparison between MALUNet and MALUNet (EED), EGE-UNet and EGE-UNet (LMD), Att-UNet and Att-UNet (EED),UNet and UNet (HD), respectively. The leftmost column represents the original image, the second column displays the ground truth, the third column shows the predicted mask for the original network, and the last column illustrates the predicted mask for the network after applying the nmODEs decoders.}
\label{fig5:env}
\end{figure*}

\subsection{Ablation Study}

In our ablation experiments using UNet as the base network on the PH2 dataset, our proposed decoders demonstrated a significant reduction in parameters. This reduction can be attributed not only to the efficient computational mechanism of NODEs but also to the fact that the number of channels in the initial $y(0)$ is only 3, maintaining consistency with the upward path. The ablation experiments involved starting with the original UNet and progressively reducing the number of channels in its upward path to 3. Subsequently, we simplified the nmODEs decoders by retaining only the part of the skip connection summation, denoted as $y^{l-1} = y^l + g(x^l,\theta^l)$, ultimately utilizing all three decoders.

\begin{table}[htbp]
  \centering
    \scalebox{0.9}{
    \begin{tabular}{ccccc}
    \toprule
    \textbf{Decoder} & \textbf{Params} & \textbf{GFLOPs} & \textbf{mIoU}  & \textbf{DSC} \\
    \midrule
    origin & 13.40M  & 31.12 & 0.8780  & 0.9342  \\
    simplified & 9.45M  & 14.92 & 0.8709 & 0.9310 \\
    simplified nmODEs & 9.42M  & 14.96 & 0.8799 & 0.9361 \\
    EED & 9.45M  & 14.89 & 0.9020  & 0.9485  \\
    LMD & 9.45M  & 14.89 & 0.9052  & 0.9503  \\
    HD & 9.48M  & 15.11 & 0.8834  & 0.9381  \\
    \bottomrule
    \end{tabular}%
    }
  \caption{Ablation experiments on network upward path structure.}
  \label{tab:addlabel2}%
\end{table}%

Analyzing the experimental results presented in Table 2 for the PH2 dataset, it is observed that the segmentation performance experiences a slight degradation when reducing the channels in the upward path of the original UNet to 3. This slight impact could be attributed to the unchanged downward path of the network, crucial for extracting feature information across different levels.  In UNet, the upward path primarily integrates information and produces output, with the final output channels consistently minimal. We hypothesize that the number of channels in the upward path of UNet might not significantly affect information integration capabilities, with the primary influence residing in the algorithm inside the decoder. Upon applying the simplified nmODEs decoder, a slight improvement in the network's performance is observed. However, substantial enhancement occurs with the complete nmODEs decoders using three different discretizations, confirming the significant influence of the decoder algorithm on information integration capabilities.

\begin{table}[htbp]
  \centering
  \scalebox{0.9}{
    \begin{tabular}{cccccc}
    \toprule
    \textbf{Decoder} & \textbf{Num.} & \textbf{Params} & \textbf{GFLOPs} & \textbf{mIoU}  & \textbf{DSC} \\
    \midrule
    \multirow{3}[2]{*}{EED} & 1     & 9.42M  & 14.75 & 0.9017  & 0.9483  \\
          & 3     & 9.45M  & 14.89 & 0.9020  & 0.9485  \\
          & 8     & 9.52M  & 15.26 & 0.9026  & 0.9488  \\
    \midrule
    \multicolumn{1}{c}{\multirow{3}[1]{*}{LMD}} & 1     & 9.42M  & 14.75 & 0.9026  & 0.9488  \\
          & 3     & 9.45M  & 14.89 & 0.9052  & 0.9503  \\
          & 8     & 9.52M  & 15.26 & 0.9050  & 0.9501  \\
          \midrule
    \multirow{3}[1]{*}{HD} & 1     & 9.43M  & 14.82 & 0.8799  & 0.9361  \\
          & 3     & 9.48M  & 15.11 & 0.8834  & 0.9381 \\
          & 8     & 9.59M  & 15.83 & 0.8917  & 0.9428  \\
    \bottomrule
    \end{tabular}%
    }
  \caption{Ablation experiments on the number of initial y-channels.}
  \label{tab:addlabel3}%
\end{table}%

The number of channels in the initial $y(0)$ plays a pivotal role in determining the overall network complexity, exerting a significant impact on its information integration capabilities. Regarded as a crucial hyperparameter, its selection is of paramount importance. Table 3 provides a thorough comparison of nmODEs decoders employing different numbers of channels in the initial $y(0)$. The table illustrates that a three-channel $y(0)$, aligning with the input information, proves more effective than a single-channel $y(0)$. Similarly, increasing the number of channels in $y(0)$ to 8 leads to an improvement in the network's performance. Opting for an even higher number of channels, such as 16 or 32, would lead to an escalation in the number of parameters and network complexity, deviating from our original intention. Therefore, such configurations are not utilized.

\section{Conclusion}
In this paper, we introduce three specialized nmODEs decoders tailored for U-like networks. By discretizing nmODEs using methods like explicit Euler's, Heun's, and linear multistep methods, we exploit the benefits of the upward path's low-channel parameterlessness and the memory efficiency of neural ordinary differential equations. 
Our decoders drastically reduce parameters and computational demands in U-like networks. To thoroughly assess their impact, we undertake a comprehensive series of segmentation experiments across widely-used datasets. We conduct an extensive comparative analysis between prevalent U-like networks and their nmODEs-enhanced counterparts. The outcomes of our study shed light on the streamlined complexity exhibited by networks that integrate nmODEs decoders, showcasing their remarkable capability to either maintain or enhance performance. Furthermore, our findings indicate that in specific tasks and network configurations, there exists a promising potential to achieve state-of-the-art performance levels. We hope that our work will provide fresh insights for the development of lightweight medical image segmentation models.

\section*{Acknowledgements}
This work was supported by the National Major Science and
Technology Projects of China under Grant 2018AAA0100201, the National Natural Science Foundation of China under Grant 62206189, and the China Postdoctoral Science Foundation under Grant 2023M732427.

\bibliographystyle{named}
\bibliography{ijcai24}

\end{document}